\def\year{2018}\relax
\documentclass[letterpaper]{article} 
\usepackage{booktabs} 
\usepackage{graphicx,balance} 
\usepackage{caption,subcaption}
\usepackage{indentfirst}
\usepackage{algorithm}
\usepackage[noend]{algpseudocode}
\usepackage{amsmath}
\usepackage{amssymb}
\usepackage{graphicx}
\usepackage{subcaption}
\usepackage{balance}

\usepackage{multirow}
\usepackage{colortbl}
\usepackage{multirow}
\usepackage{rotating}

\usepackage{color}

\usepackage{aaai18}  
\usepackage{times}  
\usepackage{helvet}  
\usepackage{courier}  
\usepackage{url}  
\usepackage{graphicx}  
\begin{document}
%
\title{Tensor Embedding: A Supervised Framework for\\
Human Behavioral Data Mining and Prediction}
\author{Homa Hosseinmardi$^{1}$, Amir Ghasemian$^{1,2}$, Shrikanth Narayanan$^{1}$, Kristina Lerman$^{1}$, Emilio Ferrara$^{1}$\\
\{homahoss,lerman,ferrarae,shri\}@isi.edu, amgh5286@colorado.edu\\
 $^{1}$University of Southern California, Information Sciences Institute, Marina Del Rey, CA\\
$^{2}$Computer Science Department,  University of Colorado Boulder, CO}


\maketitle
\begin{abstract}
Today's densely instrumented world offers tremendous opportunities for continuous acquisition and analysis of 
multimodal sensor data providing temporal characterization of an individual's behaviors. Is it possible to efficiently couple such rich sensor data with predictive modeling techniques to provide contextual, and insightful assessments of individual performance and wellbeing? Prediction of different aspects of human behavior from these noisy, incomplete, and heterogeneous bio-behavioral temporal data is a challenging problem, beyond unsupervised discovery of latent structures. We propose a Supervised Tensor Embedding (STE) algorithm for high dimension multi-modal  data with join decomposition of input and target variable. Furthermore, we show that features selection will help to reduce the contamination in the prediction and increase the performance. The efficiency of the methods was tested via two different real world datasets. 
\end{abstract}

\section{Introduction}

Rapid improvements in sensor technology have made continuous, unobtrusive sensing of individuals practical by providing temporal streams of individual physiological and psychological states, physical activity, and  social and environmental contexts
~\cite{ghandeharioun2017objective,miluzzo2008sensing,bang2008toward,aldwin2007stress,StudentLife2014,choudhury2008mobile,Wang:2018:TDD:3200905.3191775}. Such data has, in turn, created opportunities for enhanced understanding of factors contributing to mental health and wellbeing, including in the workplace.  Several past studies collected multi-modal data from individuals in real-world settings in order to infer 
psychological and health states. For example, the 10-week StudentLife study of Dartmouth undergraduate and graduate students used passive and mobile sensor data to study wellbeing, academic performance and behavioral trends \cite{StudentLife2014,Wang:2018:TDD:3200905.3191775}. SNAPSHOT, a 30-day study on MIT undergraduates, used mobile sensors and surveys to understand sleep, social interactions, affect, performance, stress and health \cite{3-2}. RealityMining, a 9-month study of 75 MIT Media Laboratory students, used mobile sensor data to track the social interactions and networking \cite{3-3}. The friends-and-families study collected data from 130 adult members of a young family community to study fitness intervention and social incentives \cite{aharony2011social}. In contrast, our work focuses on individuals in the workplace. Specifically, the present paper is based on a study that examines the complex interplay between individual differences, job performance, and well-being in jobs with varying cognitive, affective and social demands, measured both at the workplace (and to some extent, complemented, outside the workplace). More than 50 clinical and other hospital staff were instrumented and assessed with a variety of wearable and environmental sensors during their work shift for a duration of four weeks.   

Sensors data are typically collected from participants in their natural settings, continuously and over extended time periods. Therefore, the resulting data are  heterogeneous, sparse,  and high-dimensional, with many, often hand-crafted, features. 
Feature engineering, however, quickly becomes burdensome, especially when there is more than one target variable to model in human behavioral studies, e.g., five personality traits, stress, depression, performance etc. Deep learning  has recently been  successfully used for feature extraction from audio, images, social networks and other spatio-temporal data \cite{Wang:2016:SDN:2939672.2939753,jia2014caffe,le2011learning}. However, these models require many samples (instances) for training the models, which is not often feasible in some longitudinal studies. For these high dimension tasks with small number of samples, dimensionality reduction techniques, such as Partial Least Squares (PLS) regression \cite{pls} or Principal Component Analysis (PCA)~\cite{pca} are often used. While PCA finds 
a linear combination of features with highest variation, independent of the target variable, PLS considers both the independent and dependent variables by projecting them onto a low dimensional latent variable space. 
Latent 
features obtained from unsupervised tensor decomposition
~\cite{carroll1970analysis,harshman1970foundations} can be used as features with any conventional regression method to predict  target variables. However, it is unlikely that these new features will have predictive power for all target constructs of interest. 

To address this challenge, we would like to embed the data 
into a latent space using \emph{supervised decomposition} methods, which couple dependent and independent variables in the decomposition step. With the increasing demands of problems that  involve higher-order data, classification and regression methods which predict the target variable directly from N-way input have been receiving increasing attention \cite{hoff2015multilinear,Tao2007Supervised,wu2013supervised,yu2016learning,haupt2017near,eliseyev2012l1,eliseyev2013recursive,zhao2013kernelization,zhou2013tensor,hou2016common}. These methods have been widely applied in neural signal processing, image and video processing and, even in chemistry. We would like to use these methods to predict  well-being, performance and affect from multimodal, dynamic sensor data collected in real world workplace settings. 
 
To achieve this goal, we propose a method for supervised embedding, which finds the latent components combined with a feature selection step. Any regression function can be used next for the prediction task. 

\begin{figure*}[h]
    \centering
    \includegraphics[width=0.90\textwidth]{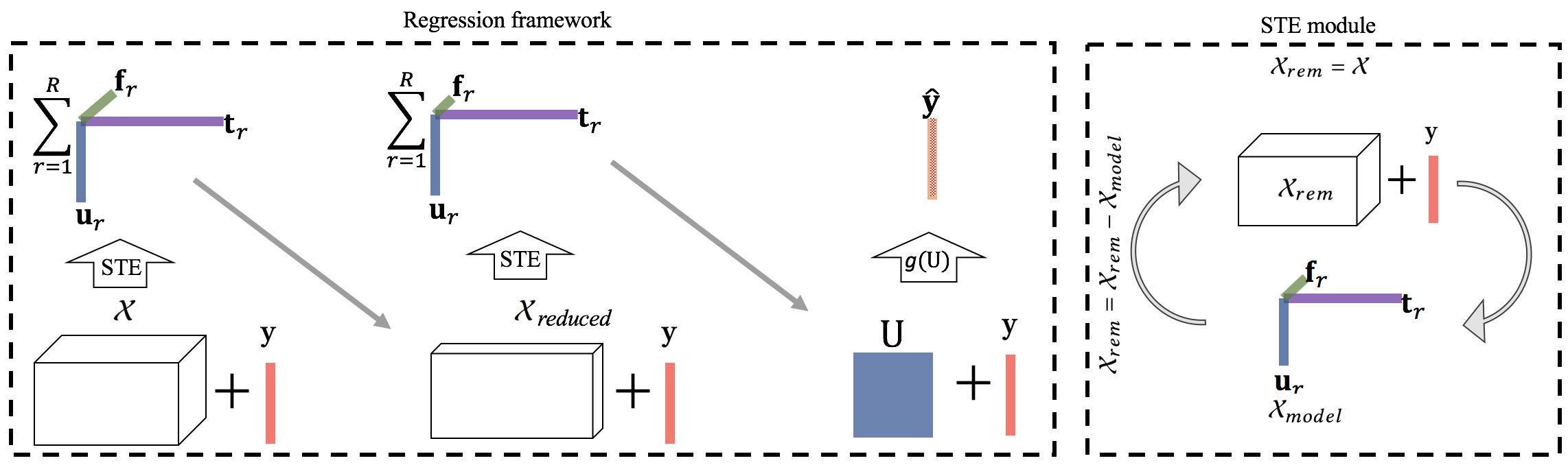}
    \caption{Left: Framework from tensor $X$ to prediction $\hat{y}$, Right: Supervised Tensor Embedding (STE) model. This process will be repeated for $r=1,2,..,R$. At each iteration the rank one model $\mathcal{X}_{model}$ will be extracted and  subtracted from the $\mathcal{X}_{rem}$.}
    \label{fig:workflow}
\end{figure*}

\subsection*{Contributions}
\begin{itemize}
  \item  A novel tensor-input/vector-output Tensor Embedding (STE) algorithm for high dimension multi-way (tensor) noisy data
  \item Variable selection using components' activation is united with supervised embedding 
  \item Validation of recovered latent patterns via prediction performance on two real world multimodal behavioral datasets including our recent ``in the wild" experimental study  that collected bio-behavioral data from subjects in challenging cognitive, social, and affective demands in their natural hospital workplace (and outside work) settings. 
\end{itemize}

\section{Related Work}
The supervised tensor learning for classification problems proposed in \cite{Tao2007Supervised} extends support vector machines (SVM) and minimax probability machines (MPM) to N-way data. Another body of work runs linear regression directly on N-way data and finds N-way coefficient tensor $\mathcal{W}$, where $\mathbf{y} = <\mathcal{X},\mathcal{W}> + \epsilon$, \cite{yu2016learning,haupt2017near,guo2012tensor}. Previously  \cite{hoff2015multilinear} has looked at tensor regression problem for relational longitudinal data. 

With a joint CP decomposition of input and target variables, NPLS is one of the widely used methods for higher order regression problems, \cite{npls}. The work in \cite{zhao2013higher} has proposed higher order partial least squares model (HOPLS) with joint orthogonal block Tucker decomposition to improve the predictability performance of NPLS by extraction of latent components based on subspace approximation rather than on low-rank approximation. However, HOPLS also performs poor under high level of noise and also is designed for target variable with number of dimensions more than one, e.g., 3D hand movement prediction. 
The recursive N-way partial least squares (RNPLS) \cite{eliseyev2013recursive} processes the tensor sequences by unifying a recursive calculation scheme with the N-way data representation of NPLS for real time applications and later recursive exponentially weighted (REW-NPLS) improves the performance of RNPLS \cite{eliseyev2017recursive}. 

In this paper we propose a supervised embedding into latent space, which finds the weights of latent components with joint CP decomposition of independent tensor and dependent target variable. While NPLS combines the projection of data in lower dimension latent space with a linear regression model, we are interested in supervised embedding which then can be combined with any regression/classification function for subsequent prediction tasks. Although the tensor decomposition will down-weight irrelevant and noisy features, discarding these features will reduce the contamination in the prediction, \cite{kuhn2013applied}. The work in \cite{eliseyev2012l1} develops an L-1 penalized NPLS algorithm, applied to sparse BCI calibration systems. Variable Importance in Projection (VIP) was proposed by  \cite{favilla2013assessing} in order to highlight the most relevant features in NPLS. Here we use the activation of features across all components after joint decomposition of dependent and independent variables as the importance measure of the features. After selecting top features, in a second iteration, we apply STE on a reduced sized tensor and regenerate the latent factors.

\section{Method for Tensor Regression}

In this section, we present our supervised learning model that is constituted by three steps: (i) supervised tensor embedding (STE), (ii) feature selection, and  (iii) regression. 
We start with some definitions and notation.

\subsection{Notation and Definitions}
A convenient mathematical representation of multi-modal data is a tensor, $\mathcal{X} \in \mathbb{R}^{I_1 \times ... \times I_P}$. The unfolded tensor in mode $d$ is defined as $X_{(d)} \in \mathbb{R}^{I_d \times I_1 ...I_{d-1}I_{d+1}... I_P}$. The d-mode vector product of a tensor $\mathcal{X}$ with vector $\mathrm{y}$ is defined as $\mathcal{Z} = \mathcal{X} \bar{\times}_1 \mathbf{y}$, where $z_{i_1 i_2... i_{d-1} i_{d+1}...i_P} = \sum_{i_d=1}^{I_d} x_{i_1i_2...i_P} y_{i_d}$. Table \ref{tab:notation} summarizes the notation used throughout this paper. 

For $N$ individuals, with $I$ time units and $J$ features, tensor $\mathcal{X} \in \mathbb{R}^{N \times I \times J} $ will be created. Entry $x_{nij}$ of this tensor corresponds to the $i^{th}$ feature of $n^{th}$ individual at the $j^{th}$ time unit. 
The covariance matrix Z is defined as Z $= <\mathcal{X},\mathbf{y}> = \mathcal{X} \bar{\times}_1 \mathbf{y}$, where $z_{ij} = \sum_{n=1}^{N} x_{nij} y_{n}$. 

The CP decomposition, will decompose the tensor $\mathcal{X} \in R^{N \times I \times J}$ into sum of rank-one tensors, called components: $\mathcal{X} =  \sum_{r=1}^{R} \lambda_r \mathbf{u}_r \circ \mathbf{t}_r \circ \mathbf{f}_r $, where $\lambda_r$ are the values of the tensor core $L = diag(\Lambda)$, and the outer product $\mathbf{u}_r~\circ~\mathbf{t}_r~\circ~\mathbf{f}_r$ corresponds to the $r^{th}$ component of rank-R estimation.



\begin{table}
	\centering
    \caption{Table of Symbols.}
	\begin{tabular}{c|c}
	\hline \hline
	Symbol & Definition \\ \hline
	\hline
    $\mathcal{X},\mathrm{X},\mathbf{x},\mathrm{x}$ & Tensor, matrix, column vector, scaler\\ \hline
	$ \mathbf{x} \in \mathbb{R}^{I} $ & Definition of an I-dimensional vector  \\	\hline
     $\circ$  & Outer product \\\hline
      $\otimes$  &  Kronecker product \\\hline
     
    $\mathcal{X} \bar{\times}_1 \mathbf{y}$ & d-mode vector product \\ \hline \hline
	\end{tabular}
	
	\label{tab:notation}  
\end{table}

\subsubsection{Supervised Tensor Embedding}\label{STE}

We are interested in finding the latent user factors of the tensor data, such that they can be of good predictive ability of the target variables of interest. Decomposition of the collected data in an unsupervised way can help find underlying structure, however these latent user factors may not necessarily have high correlation with all different human behavior aspects of interest and may only explain a subset. By applying supervised decomposition for each target variable, we would like to find the latent factors which correlate with it the most. Then we can use any regression function $g(.)$ on the obtained user latent matrix U to estimate $\hat{y}$. 

Our work builds upon the idea of N-way PLS \cite{npls}, where the algorithm constructs a model of both $\mathcal{X}$ and $\mathbf{y}$ for each component and then the models are subtracted from both $\mathcal{X}$ and $\mathbf{y}$ iteratively. We are interested in extracting rank one models from $\mathcal{X}$ iteratively, and finding the latent factors highly correlated with $\mathbf{y}$, without fitting a prediction model. For this purpose we start with building the cross-covariance matrix Z, (line 7, \ref{alg1}). The goal is finding $\mathbf{t}$ and $\mathbf{f}$ such that $\mathbf{u}$ has maximum correlation with $\mathbf{y}$. It is equivalent to solving $\max_{\mathbf{t},\mathbf{f}} \sum_{i=1}^{I} \sum_{j=1}^{J} z^2_{ij} t_i f_j $ and the answer for this problem is the first set of normalized vectors from a singular value decomposition on Z (line 8, \ref{alg1}), \cite{npls}. When $\mathbf{t}$ and $\mathbf{f}$ are extracted, we can find $\mathbf{u} = \mathcal{X} \times_1 (\mathbf{t} \otimes \mathbf{f})$. From the three latent factors, we can reconstruct the rank-1 model $\mathcal{X}_{model}$. 
 This process will be repeated $R$ times, at each iteration a rank-1 component will be extracted from data and $\mathcal{X}_{rem}$ will be created, Fig. \ref{fig:workflow}-right. After finding the latent factors U, F and T, we can look at the importance of the features and discard the irrelevant features, then again we find the latent factors on the reduced dataset.  

\begin{algorithm}
\caption{Supervised Tensor Embedding}
\label{alg1}
\begin{algorithmic}[1]
\State input: $\mathcal{X}$ independent tensor, $y$ dependent variable
\State parameters: $R$ - decomposition rank
\State output: U, T, F  
\State Center $\mathcal{X}$ and $\mathbf{y}$
\State ${\mathcal{X}}_{rem} = \mathcal{X}$
\For{r = 1:R}
\State $\mathbf{t}_r,\mathbf{f}_r \leftarrow $SVD(Z) 
\State $\mathbf{t}_r \leftarrow \mathbf{t}_r/||\mathbf{t}_r||,\mathbf{f}_r \leftarrow \mathbf{f}_r/||\mathbf{f}_r||$ 
\State $\mathbf{u}_r \leftarrow \mathcal{X}_{rem} \bar{\times}_1 (\mathbf{t}_r \otimes \mathbf{f}_r)$ 
\State $\mathcal{X}_{model} = \mathbf{u}_r \circ \mathbf{t}_r \circ \mathbf{f}_r$ 
\State $\mathcal{X}_{rem} = \mathcal{X}_{rem} - \mathcal{X}_{model}$ 
\EndFor
\end{algorithmic}
\end{algorithm}

\subsubsection{Feature Selection}
Given different desired target variables for the same input, not all the features are equally informative for the different targets. Therefore, a proper feature selection can improve predictability of the reduced sized input data. Although the STE model down weights the irrelevant features, it does not discard them. As a result, it is possible that a large number of irrelevant features can still contaminate the predictions. We use feature activation in the latent factors and drop the uninformative features.

 The feature activation vector defines the importance of each feature for the specific prediction task. We use the latent factor matrix F and extract the feature importance as following:
\vspace{-3mm}
\begin{equation*}
FI_i = \sum_{r=1}^{R} f^2_{ri},~~~~ i=1,2,...,J
\end{equation*}

Given the feature importance score, any desired technique can be applied for feature selection. In this paper we have chosen filter methods and we pick top $K$ features and discard the others.

\begin{figure*}[!htb]
    \centering
    \includegraphics[width=0.95\textwidth]{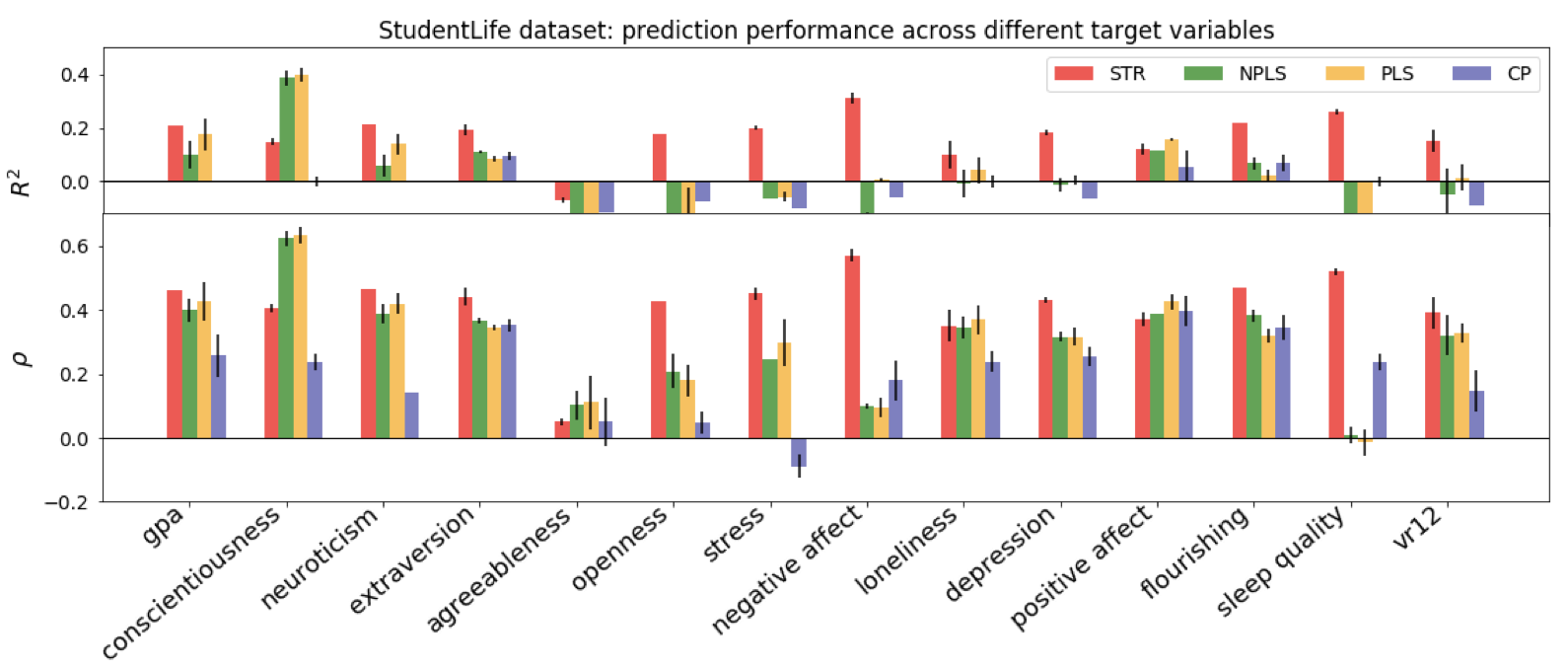}
    \vspace{-4mm}
    \caption{Prediction performance (top:$R^2$, bottom: pearson correlation $\rho$) across different target variables, using STR, NPLS, PLS, CP methods.}
    \label{fig:SL}
\end{figure*}

\subsubsection{Regression model}
Previously we introduced our supervised decomposition algorithm to obtain the user latent factors with high correlation with target variable of interest and then defined feature importance to discard irrelevant features. The latent factors can be used for exploration of active features, temporal trends and similar users given a certain target variable. Furthermore, we can apply any regression function $g(.)$ for inference of dependent variable. The parameter of our model would be number of features to keep, number of components, and  parameters of the regression function $g(.)$, algorithm \ref{alg2}.

\begin{algorithm}
\caption{Supervised Tensor Regression}\label{alg2}
\begin{algorithmic}[1]
\State input: $\mathcal{X}$ independent tensor, $y$ dependent variable
\State parameters: $K$ - number of features, $R$ - decomposition rank , $g(.)$- regression model
\State output: $\hat{y}$  
\State Compute component activation for each sensory variable using STE
\State Form a reduced data tensor $\mathcal{X}_{reduced}$ consisting of only time series whose activation power is among top K variables
\State Compute latent variables using STE 
\State Use user latent variables in a regression model to predict outcome $\hat{y} = g(U)$
\State Pick K and parameter of model $g(.)$ by cross validation 
\end{algorithmic}
\end{algorithm}

\section{Results}
Our goal is to understand whether supervised decomposition can find the low-dimensional structure of daily life from wearable devices that better  correlates with target behavioral constructs. We test our models along with CP, NPLS and PLS on two real world datasets described below. STR, CP, and NPLS were used to model the data in a tensor form, where PLS was used on a mode-1 matricized version of the same tensor. To compare the predictability, we compare coefficient of determination $R^2$ and pearson correlation $\rho$ obtained from each method. Because we have a small number of samples, we present the results on test set in a nested cross-validation, when in the train-validation set we tune the parameter of the model and then we present the result on test set. For STR method, we test different regression functions , with and without feature selection and we present the best result. Later in Figs. \ref{meshplots_studentlife} and \ref{svr_ridge}, we investigate the effect of feature selection and choice of regression function.  We have repeated each experiment 20 times and have reported mean value for $R^2$ and $\rho$. In the bar graphs standard deviation is also reported.

\subsection{StudentLife Data}
StudentLife is a 10-week study conducted during 2013 spring semester on 48 Dartmouth students (30 undergraduate and 18 graduate students), \cite{StudentLife2014}. Psychometric data were collected from student via a pre-Assessment and post-assessment Survey. GPA was also collected at the end of the semester, which will be used as a measure of students academic performance. The other surveys include Big Five Inventory (BFI), Positive Affect and Negative Affect Schedule (PANAS), Perceived Stress scale \cite{cohen1983global}, UCLA loneliness scale \cite{russell1996ucla}, (PHQ9) \cite{kroenke2002phq}, flourishing scale \cite{diener2010new}, (VR12) \cite{vr12} measuring students of wellbeing, and Pittsburgh Sleep Quality Index (PSQI) \cite{b27} as a measure of health. 

\begin{figure*}[!htb]
    \centering
    \includegraphics[width=0.95\textwidth]{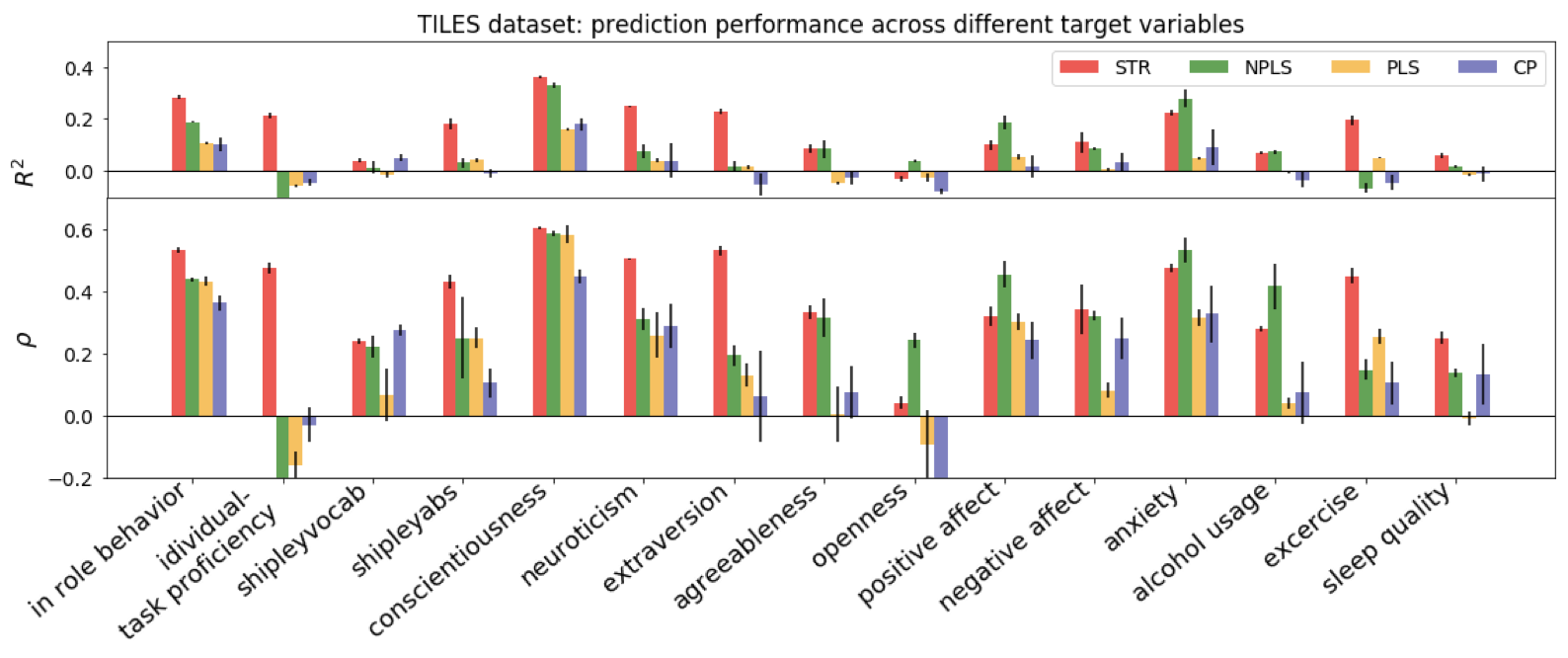}
    \vspace{-4mm}
    \caption{Prediction performance (top:$R^2$, bottom: pearson correlation $\rho$) across different target variables, using STR, NPLS, PLS and CP.}
    \label{fig:tiles_bar}
\end{figure*}

Using the raw sensor data collected from students, physical activity (stationary, walk, run and unknown) and audio activity (silence, voice, noise and unknown) have been inferred. To create our tensor, each time unit comprises one day worth of data, and is divided into four time bins, bedtime (midnight-6 am), morning (6 am-12 pm), afternoon (12 pm-6 pm), and evening (6 pm-midnight). We extract duration (minutes) of running, walking, stationary, silence, voice, noise, and dark, per time-bin in each day. Frequency and number of changes in each behavior (e.g. from walking to running) for each time-bin has been also captured. From GPS and WiFi, the number of unique locations visited, and from Bluetooth, the number of unique nearby devices per time-bin are added to the variable set. We normalize all the variables across time dimension to have the same range $[0,1]$ to avoid variables with large values (e.g. duration in minutes) dominate the analysis. At the end, we organize our data as tensor $\mathcal{X}$ with $N=46$ individuals, $I=108$ features and $K=63$ days. Only 5\% of the tensor is missing, which we imputed by filling them with the mean value. Numbers of samples for different targets varies from 30 to 46, as not every participant had answered all the surveys.  

Figure \ref{fig:SL} presents the results for StudentLife dataset. In Figure \ref{fig:SL} we can see that $R^2$ has improved significantly for some of the target variables, e.g. negative effect, flourishing and sleep quality. Also it is worth mentioning that we are using only passively collected data for all prediction tasks, without using any EMA or self-report values by the participants as the features. 

\subsection{TILES Data} \label{tiles_data}
Tracking Individual Performance with Sensors ({TILES}) study, is an ongoing research study of workplace performance which measures physical activity, social interaction, physiological state of employees (such as nurses) in a large university hospital setting.  The study aims to collect data from more than 300 participants over a 10-week period during the spring, summer and fall of 2018. It includes sensor and health data, psychometrics and job performance measures. Sensor data were collected from garment-based  wearable sensors (OmSignal) and  wristbands (Fitbit). OmSignal is a Biometric Smartwear company that produces smart under shirts and bras. 
Their garments include health sensors embedded into the fabric that measure biometric data in real-time and can relay this information to the participant's smartphone. OMsignal sensor provides information such as heart rate, heart rate variability (HRV), breathing, and accelerometery (to provide sitting position, foot movement and more). 
Fitbit collects heart rate, steps, sleep and cardio information. 
Participants were asked to wear their Fitbit 24/7. However, they were instructed to wear OmSignal sensors only during their work shifts. It is worth mentioning that clinical staff in this study work a minimum of 3 days per week (in 12 hour shifts), which can be any day during weekdays or weekend. Also some belong to day shift and others to night-shift, which would be 7am-7pm or 7pm-7am, respectively. 

Psychometric data were collected from participants via pre- and post study surveys. These surveys measured job performance, cognitive ability, personality, affect, and health state and are used as our groundtruth in the models. More specifically, the target variables we predict in this dataset include the In-Role Behavior Scale \cite{b16}, Individual Task Proficiency Scale \cite{b17}, Shipley 2 \cite{b20}, Big Five Inventory (BFI) \cite{b21}, Positive Affect and Negative Affect Schedule (PANAS) \cite{b22}, State-Trait Anxiety Inventory (STAI) \cite{b23}, Alcohol use Disorders Identification Test (AUDIT) \cite{b24}, International Physical Activity Questionnaire (IPAQ) \cite{b26}, and Pittsburgh Sleep Quality Index (PSQI) \cite{b27}. 

  For the experiments reported in this paper, we use the pre-survey scores provided by the participants (clinical staff in a large hospital) as the target variables for TILES dataset. Similar to the previous dataset, each time unit comprises one day's worth of data, and is divided into four time bins: bedtime (midnight-6 am), morning (6 am-12 pm), afternoon (12 pm-6 pm), and evening (6 pm-midnight). We extract a set of statistics such as mean, standard deviation, kurtosis, etc. from each time series from the OMSignal and Fitbit sensor streams, in each time-bin. TILES is an ongoing project and we have data for about 50 participants in the first wave of the data collected in spring which we use in this paper. We organize the data as tensor $\mathcal{X}$ with $I=50$ individuals, $J=1225$ features and $K=30$ days. About 60\% of the tensor is missing, which we imputed by filling them with the mean value for each time series. 
We tested our model for 15 different target variables in comparison with PLS, NPLS and CP plus a regression function. Similar to StudentLife dataset, we divided the data into train, validation and test set, and performed nested cross validation and report the result on test set, Fig. \ref{fig:tiles_bar}.

\subsection{Analysis}
Looking at 29 different predictions across two datasets, STR outperforms the other three methods in 21 tasks. 
Some of the constructs were not predictable with any of approaches which can be due to the lack of appropriate features, or inadequate feature engineering. Also it is possible tensor tri-linear models are not suitable for modeling those constructs, as by applying the no-free-lunch idea to all sorts of scientific problems it has been shown that different type of algorithms may work well for different type of problems, \cite{munoz2017instance}. The improvement in performance by STR comes from the partial contribution of 1) change in the deflation of $y$, 2) choice of regression function or 3) feature selection. In NPLS model, at each iteration, the estimated part of $y$ will be subtracted from it. By not subtracting the explained variation from $y$ every iteration, the cost would be higher correlation among the features. However, it will not come at the cost of less accurate performance for all constructs. In order to understand the effect of deflation of y, we use a simple ordinary least square model as regression function of STR with no feature selection on the time series and compare it with NPLS. For 10 target variables out of 29 total, both $R^2$ and $\rho$ improved. As another contribution, by separating the embedding and regression steps in STR model, the latent features can be tested and paired with the most appropriate regression function to improve the performance. For example, for negative affect from StudentLife dataset and shipley abstract from TILES, we have applied Ridge and linear SVM as two different regression models, which we can see SVR will lead to 4\% improvement in both $R^2$ and $\rho$, Fig. \ref{svr_ridge}.

\begin{figure}[!htb]
  \centering
  \begin{tabular}{|*{8}{c|}}
    \hline
    \multicolumn{1}{|>{}c|}{} & \multicolumn{2}{c|}{negative affect} & \multicolumn{2}{c|}{shipely abs} \\
    \arrayrulecolor{black}
    \cline{2-5}
    \multicolumn{1}{|>{}c|}{} & SVR & Ridge & SVR & Ridge \\
    \hline

     r2 & 0.31 & 0.27 & 0.19 & 0.15  \\ \cline{1-4}
     \hline
     ro & 0.57 & 0.53 & 0.44 & 0.39  \\ \cline{1-4}
    \hline
  \end{tabular}
  \vspace{-1mm}
  \caption{The effect of using different regression models is present by two examples; StudentLife: negative affect, TILES: shipley abs}
  \label{svr_ridge}
\end{figure}

\begin{figure}[h]
\centering
\begin{subfigure}{.25\textwidth}
    \centering
    \includegraphics[width=0.9\textwidth]{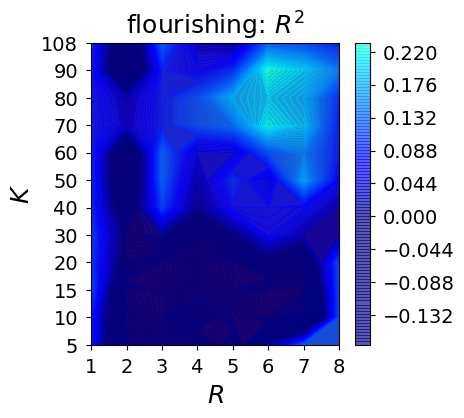}
\end{subfigure}%
\begin{subfigure}{.25\textwidth}
    \centering
    \includegraphics[width=0.9\textwidth]{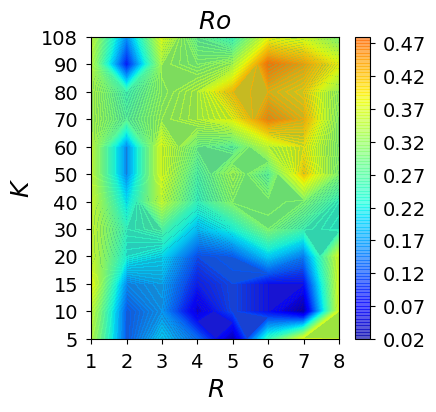}
\end{subfigure}
\begin{subfigure}{.25\textwidth}
    \centering
    \includegraphics[width=0.9\textwidth]{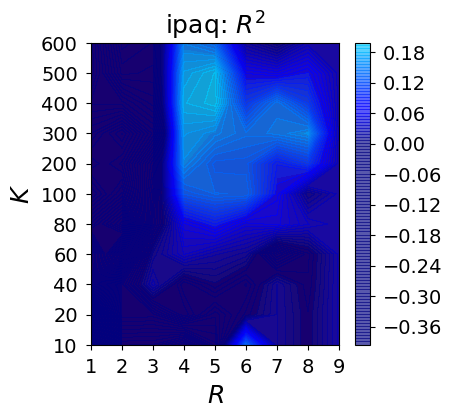}
\end{subfigure}%
\begin{subfigure}{.25\textwidth}
    \centering
    \includegraphics[width=0.9\textwidth]{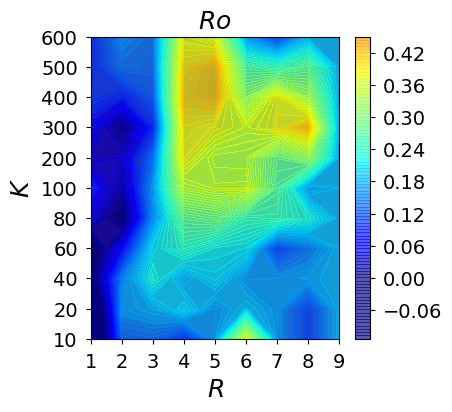}
\end{subfigure}
\vspace{-3mm}
\caption{Prediction performance (left: average $R^2$, right: average pearson correlation)  versus number of latent factors (R) and number of selected features (K) for StudentLife: flourishing  (top) and TILES: negative affect (bottom).}
\label{meshplots_studentlife}
\end{figure}

Parameter selection effect has been demonstrated in Fig. \ref{meshplots_studentlife}, for $R^2$ and $\rho$ versus number of components and number of selected features. The best result for negative effect from StudenLife dataset is obtained at $K=50$ and $R=4$. Also we can see the effect of number of features $K$ and model rank $R$, Fig. \ref{meshplots_studentlife}.

\subsection{Case Study: Student Performance}
\begin{figure}[h]
\centering
\begin{subfigure}{.5\textwidth}
    \centering
    \includegraphics[width=0.8\textwidth]{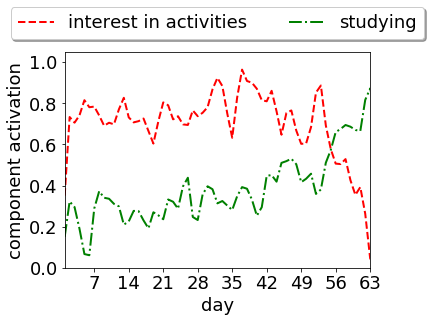}
\end{subfigure}%
\vspace{-4mm}
\caption{Temporal pattern of first temporal components associated with GPA and first temporal components associated with depression.}
\label{tempral_pattern_gpa}
\end{figure}

In this section we look at the temporal patterns obtained from different supervised decomposition tasks. One interesting observation is that GPA's first latent temporal factor increases towards end of the semester and depression's first latent temporal factor start decreasing around after mid-semester. The duration of being physically stationary in the afternoon, evening and midnight, the duration of audio silence, and the number of detected on-campus wifi locations are among the top features of the first 
latent factor of GPA. This inferred latent component can be an indicator of studying in a quiet environment (the studying factor). To obtain a better understanding of the correlation between depression and performance, we first looked at depression latent factor. The top activated features include running, walking during evening, duration of conversation  in the morning, and the number of on-campus wifi locations. This latent behavior has features related to school engagement activity. We name it as "diminished interest in activities", as the temporal trend decreases over the second half of the semester, Fig. \ref{tempral_pattern_gpa}, green graph. For the students that have higher value in this latent factor, there was a higher chance of depression. The user latent factor of "diminished interest in activities" has a correlation of -0.9 with the studying user latent factor. It can mean that students who grow depressive symptoms over the semester have lower performance at the end of the semester. Recently in another study \cite{Wang:2018:TDD:3200905.3191775}, it has been observed that depression has negative correlation with the slope of the duration of time students spent in study places during the semester on-campus.

\section{Conclusions and Future work}

Rich multimodal data collected from wearable sensors (e.g. Fitbit), mobile phones, online social networks, etc is  becoming increasingly available to reconstruct digital trails and study human behavior. In this paper, we use two real world datasets---TILES and StudentLife---collected using passive and mobile sensors, with the goal of inferring wellbeing, performance, and personality traits. We developed a learning framework based on \textit{supervised tensor embedding} to find latent space that is highly correlated with target variables of interest. This type of decomposition can uncover latent user factors which are strong predictors of target variables. Further we explore how variable selection can improve the prediction performance and propose a robust variable selection frame work. One limitation of our work is that the framework captures only linear structure. Another limitation in using prediction performance as a metric for selection of best rank and $K$ (number of top features). We plan to use kernel methods for nonlinear projection and defining information theoretic metrics for best embedding. Feature selection can be extended to be applied on latent features too. Also, as TILES study is an ongoing project, we plan to implement supervised predictions of individuals' performance and personality directly from different modalities, such as social media activity, location, audio.

\bibliographystyle{aaai}
\balance
\bibliography{bibliography} 

\end{document}